\documentclass[preprint,12pt]{elsarticle}

\usepackage{subfigure}
\usepackage{amssymb}
\usepackage{graphicx}
\usepackage{algorithm}
\usepackage{algorithmic}
\usepackage{url}
\usepackage[T1]{fontenc}
\usepackage[latin9]{inputenc}
\usepackage{graphicx}
\usepackage{setspace}

\journal{Elsevier}

\begin{document}

\begin{frontmatter}



\title{Feature Selection Based on Confidence Machine}


\author{Chang~Liu and Yi~Xu}

\address{Department of Computer Science,
Montana State University,
Bozeman, MT 59717.
e-mail: chang.liu@cs.montana.edu; xuyi0409@gmail.com\\
}

\begin{abstract}
In machine learning and pattern recognition, feature selection has
been a hot topic in the literature. Unsupervised feature selection
is challenging due to the loss of labels which would supply the related
information.How to define an appropriate metric is the key for feature
selection. We propose a filter method for unsupervised feature selection
which is based on the \textquotedblleft{}Confidence Machine\textquotedblright{}.
Confidence Machine offers an estimation of confidence on a feature\textquoteright{}s
\textquotedblleft{}reliability\textquotedblright{}. In this paper,
we provide the math model of Confidence Machine in the context of
feature selection, which maximizes the relevance and minimizes the redundancy
of the selected feature. We compare our method against classic feature
selection methods Laplacian Score, Pearson Correlation and Principal
Component Analysis on benchmark data sets. The experimental results demonstrate
the efficiency and effectiveness of our method.
\end{abstract}


\end{frontmatter}


\section{Introduction}
Feature selection is a key technology to deal with high-dimensional
data in machine learning \cite{jour8}. It is reported from 2012 the big data "size" is moving from a few dozen terabytes to many petabytes. How to deal with the huge
data in case of \textquotedblleft{}curse of dimensionality\textquotedblright{}
is essential in real applications \cite{jour42}. Feature selection, as a powerful
method of dimension reduction \cite{jinadd6}, has been successfully applied in pattern
recognition {[}1{]}, computer version {[}2{]}, \cite{proceeding1}, active learning \cite{jour501}, and sparse coding \cite{jour51}, \cite{jour1}. Functionally, feature selection \cite{jinadd4}, \cite{jinadd5} is divided into three groups: filter model, wrapper model and
embedded model. Filter is the most popular model in recent research,
as it has low computational cost and is robust in theoretical analysis. Depending
on the class labels, feature selection is implemented in supervised fashion or
unsupervised fashion. Most existing filter models are supervised. In real applications, the class labels are always scarce {[}3{]} \cite{jinadd8}.
It is meaningful to design a filter feature selection method in unsupervised
fashion. The criteria of maximum dependency has been studied widely
in the field of feature selection: selecting the features with highest
relevance to the target class C and at the same time minimizing the
redundancy with the rest of the features. This criteria is met in the fashions
of mutual information or correlation. In this paper, we utilize the
correlation to compute the distance among selected features, the target
class and the rest of the features. Given the feature\textquoteright{}s relevance
and redundancy, which are denoted by correlation scores, we evaluate
these properties and build a mathematic model of Confidence Machine
as the primary part of a new unsupervised filter method for feature
selection.

The main contributions of the paper are summarized as follows: a
new feature selection filter model is proposed, based on the idea
of Confidence Machine and correlation. Based on the model of Confidence
Machine, a feature\textquoteright{}s relevance and redundancy with
the rest features and with the target class are calculated, in the
way of correlation. The relevance is maximized and redundancy is minimized.
The proposed method is applied to UCI {[}4{]} benchmark data sets
(dual category and multiple category). A 2-D visualization case study
is carried out and compared with classic filter feature selection methods
(Principal Component Analysis {[}5{]}, Laplacian Score{[}6{]} and
Pearson Correlation {[}5{]}). Then comparison experiments of feature-based classifications are conducted to demonstrate the efficiency and effectiveness of our method.

\section{\emph{Feature score based on Confidence Machine}}

Before introducing the idea of Confidence Machine, let\textquoteright{}s
first talk about a wide spread consensus \textquotedblleft{}Max-Relevance\textquotedblright{}
and \textquotedblleft{}Min-Redundancy\textquotedblright{}, which was
firstly introduced by Hanchuan Peng in{[}7{]}. In feature selection,
it has been recognized that an optimal feature often means minimal
classification error. That requires the maximal statistical dependency
of the target class \emph{C} on the data distribution in the subspace.
This scheme is called maximal dependency. However, combinations of
individually good features do not necessarily lead to good classification
performance. Redundant variables within a subspace should be removed
and this property is called minimal redundancy. Usually the relevance
and redundancy is characterized in terms of correlation or mutual
information. In this paper, we choose to use correlation. In the following
part we will discuss the definition and formal math model of confidence
machine.

The algorithm of Confidence Machine describes a measure of \textquotedblleft{}reliability\textquotedblright{}
for every prediction made, in contrast to the algorithms that output
\textquotedblleft{}bare\textquotedblright{} predictions only. The
estimation of prediction confidence is represented by the P-value,
which is an indication of how good the selected feature is, in the
context of feature selection. The general idea is that the confidence
P-value corresponds to the certainty of a feature being the right
choice. If a feature has a bigger P-value, then it means this feature
has a greater relevance to the target class and at the same time
has a smaller redundancy with other features.

Now, we give the formal mathematical model of Confidence Machine in
the context of feature selection. Imagine we have a data set with
\emph{n+1} dimensions \emph{Y=$\left\{ y_{1},y_{2},\ldots,y_{i},\ldots,y_{n},C\right\} $},
in which the first $n$ dimensions are data features and the last dimension
is the class label. The ideal feature among $n$ dimensions has the property
of \textquotedblleft{}Maximum relevance\textquotedblright{} and \textquotedblleft{}Minimum
redundancy\textquotedblright{}. For each feature, we need to calculate
its relevance score \emph{pl }and redundancy score \emph{ps}.

$Pl_{i}$ is defined as the correlation between the current feature
$y_{i}$ and the target class $C$. According to the definition
of Pearson\textquoteright{}s correlation, which is the most familiar
measure of dependence between two quantities, the relevance value
of feature $y_{i}$ is defined as:

\begin{center}
\emph{$Pl_{i}$=$P{}_{y_{i},C}$=$\frac{cov(y_{i},C)}{\sigma_{y_{i}}\sigma_{C}}$=$\frac{E\left[\left(y_{i}-\mu_{y_{i}}\right)\left(C-\mu_{C}\right)\right]}{\sigma_{y_{i}}\sigma_{C}}$}
\par\end{center}

\emph{Ps$i$} is defined as the correlation between the selected feature
yi and other features. It is calculated as:

\begin{center}
\emph{Ps$_{i}$=$\sum\left(|Ps_{i1}|,|Ps_{i2}|,\ldots,|Ps_{in}|\right)$}
\par\end{center}

And \emph{Ps$ij$(j=1,2,..,n) }is the correlation value between feature
\emph{i} and \emph{j}:

\begin{center}
\emph{$P{}_{yi,yj}$=$\frac{cov(y_{i},y_{j})}{\sigma_{y_{i}}\sigma_{y_{j}}}$=$\frac{E\left[\left(y_{i}-\mu_{y_{i}}\right)\left(C-\mu_{y_{j}}\right)\right]}{\sigma_{y_{i}}\sigma_{y_{j}}}$}
\par\end{center}

For each feature \emph{y$i$}, relevance \emph{Pl$i$} is the value
needs to be maximized and redundancy \emph{Ps$i$} is the value needs
to be minimized, if \emph{y$i$} is an optimal choice. Then non-conforming
score \emph{$\mathcal{\alpha}$} is introduced. This measure directly
defines the relevance and redundancy of the current feature in relation
to the rest features and the class label. In our case the non-conforming
score for a feature \emph{i} is defined as:

\begin{doublespace}
\begin{center}
$\alpha_{i}$=$\frac{Pli}{Psi}$
\par\end{center}
\end{doublespace}

\emph{Ps$i$} is safe as a denominator, since it will always greater
than 0. And that is the reason why \emph{Ps$i$} is calculated as
the sum of absolute value of \emph{Ps$ij$(j=1,2,\ldots{},n)}. This
is a natural measure to use, as the non-conformity of a feature increases
when the distance from the class becomes bigger or when the distance
from the other features becomes smaller.

Provided with the definition of non-conforming score, we will use
the following formula to compute the p-value:

\begin{singlespace}
\begin{center}
\emph{P$(\alpha_{now})$=$\frac{\#\left\{ i:\alpha_{i}>\alpha_{now}\right\} }{n}$}
\par\end{center}
\end{singlespace}

In the above equation, \# denotes the cardinality of the set, which
is computed as the number of elements in finite set. \emph{$\mathcal{\alpha}_{i}$}
is the non-conforming score of the test feature. From the above discussion,
we can see that the algorithm will output a sequence \emph{$\left\{ \alpha_{1},\alpha_{2},\ldots,\alpha_{n}\right\} $},
and based on these non-conforming scores a sequence of confidence
value \emph{$\left\{ P_{1},P_{2},\ldots,P_{n}\right\} $} will be
produced. Every P-value denotes the reliability of a feature
and is counted as the score of that feature. At the end of our algorithm,
those features with high confidence (P-value) will be chosen.

\section{\emph{Experimental evaluation}}

In this section, the empirical experiments are conducted on ten data
sets from UCI Repository {[}4{]} to demonstrate the effectiveness of
our method. There are six binary data sets and four multiple categorical
data sets. The detailed information for the data sets are listed in Table
1. In the experiment, each data set is randomly separated into two
equal parts. One part is training data and the rest of the parts are testing
data. We used the training data to build the model for feature selection.
Four filter feature selection models are utilized for comparison experiments:
Our proposed unsupervised Filter model via Confidence Machine, Pearson
correlation, Laplacian score and Principal Component Analysis. We use
Liu\_Corr, PER, LAP and PCA as abbreviations to denote these four methods
in the experiment.

\begin{table}
  \centering
  \includegraphics[width=120mm]{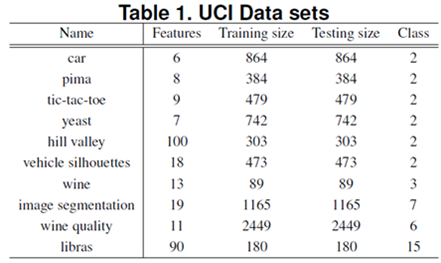}\\
  \caption{\emph{UCI data sets}}
\end{table}

\subsection{\emph{Case study of 2-D visualization}}

A simple case study for dataset \emph{wine} is shown. In total, there are 13 features
for \emph{wine}, such as \textquotedblleft{}Alcohol\textquotedblright{},\textquotedblleft{}Magnesium\textquotedblright{}
and \textquotedblleft{}Proline\textquotedblright{}. We use four filter
methods based on training data (with size 89) and apply on the testing
data. Each method chooses two features for 2-D visualization on testing
data. The results are shown in Fig.1. Two features are selected by
4 different methods and plotted in each sub figure. It can be observed
that the feature \textquotedblleft{}Flavanoids\textquotedblright{}
and feature \textquotedblleft{}Color intensity\textquotedblright{}selected
by Liu\_Corr method are crucial for discrimination.

\begin{figure}
  \centering
  \includegraphics[width=140mm]{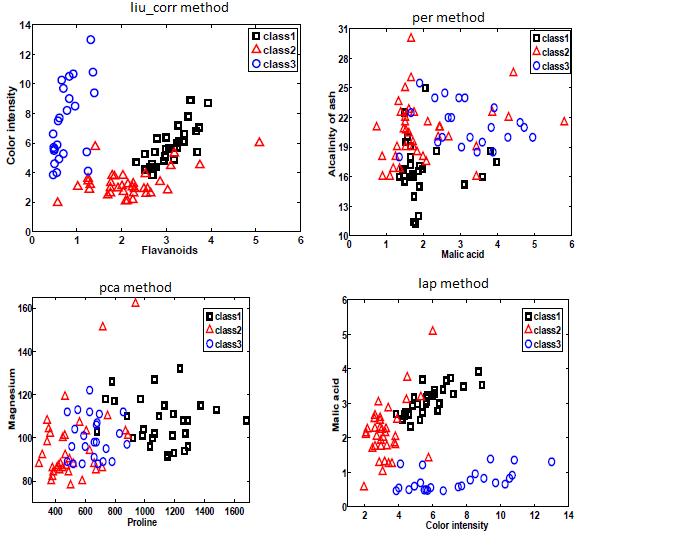}\\
  \caption{\emph{Data wine plotted in 2-D with selected features. 4 methods have
selected different 2 features}}
\end{figure}

\subsection{\emph{Feature based classification}}

When selected features are more than two, we used the feature based
classification to compare the feature selection methods. The experiment
is conducted five times and mean outputs are obtained. The target selected
features size is from one to around 80\% of whole feature size to give
comprehensive comparison. In order to show the classification performances,
we use three classic classifiers: k nearest neighbor (k = 5 in the
experiment) and LibSVM. For brevity, we only plot one data set (a
multi-category data set) result. We abbreviate the classifiers as LibSVM
and KNN in the figures.Fig.2 shows the comparison results for data
winequality. When the selected features size is greater than 4,our
method results rank first with LibSVM classifier. And our method
ranks second when the selected feature size is smaller than 4. In the
case of classifier KNN in Fig.3, the performances of Liu\_Corr rank
first when we select 6 features, and rank second with other feature
sizes. It is important to note our method is competitive to Pearson\textquoteright{}s
correlation in most feature sizes. In order to give intensive comparison
of different feature selection methods on multiple data sets, the
mean accuracy in low dimension (from feature size one to around 40\%
of whole feature sizes) are calculated based on each data set and
each classifier. Table 2 shows the detail mean outputs and the comparison
results. The highest accuracy for each classifier is highlighted. It
can be observed that our filter method won 6 and 4 times of 10 data sets
with classifiers LibSVM and KNN separately.

\begin{figure}
  \centering
  \includegraphics[width=140mm]{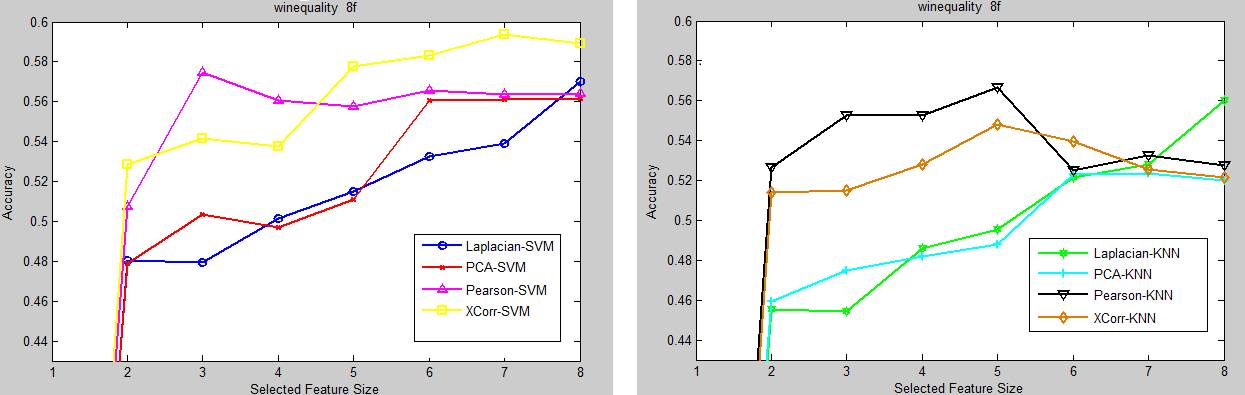}\\
  \caption{\emph{Comparison of feature based classification accuracies for data
set wine quality}}
\end{figure}

\begin{table}
  \centering
  \includegraphics[width=140mm]{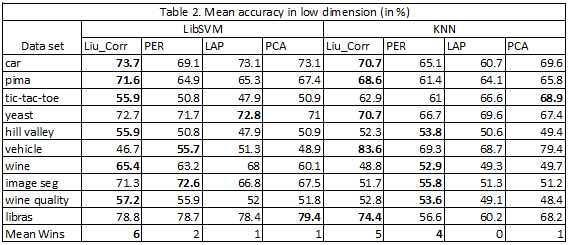}\\
  \caption{\emph{Mean accuracy in low dimension (in \%)}}
\end{table}

\section{\emph{Conclusion and future work}}

We present a new filter feature selection method in the unsupervised
fashion. Our approach aims to use correlation to evaluate the distance
between current feature with the target class, and between current
feature with other features. Then by calculating the prediction confidence
of every feature, we select the features which have the highest relevance
with class label and have the lowest redundancy with other features.
Experimental comparisons with related filter methods have demonstrated
that our method is effective in terms of visualization and classification.
Future research work will focus on increasing dimension of the data
set, statistical analysis among different filter models and improving
the theoretical framework of Confidence Machine based filter for feature
selection. Also, we plan to apply the our method to human group recognition \cite{jinadd9}, social networks analysis \cite{jinadd8} and sparse representations \cite{jour2}, \cite{jour5}.

\end{document}